\documentclass[runningheads]{llncs}
\usepackage[T1]{fontenc}
\usepackage{hyperref}
\usepackage{graphicx}
\usepackage{amsmath}
\usepackage{amssymb}
\usepackage{wrapfig}
\usepackage{multirow}
\usepackage{booktabs}
\usepackage{xcolor}
\usepackage{here}

\newcommand{\ud}[0]{\textcolor{white}{$\blacktriangle{} $}}
\newcommand{\up}[0]{$\blacktriangle{} $}
\newcommand{\dn}[0]{$\blacktriangledown{} $}

\begin{document}
\title{Experiments in News Bias Detection with Pre-Trained Neural Transformers\thanks{The authors gratefully acknowledge the funding provided by the Free State of Bavaria under its ``Hitech Agenda Bavaria''. We would also like to thank Michael Reiche for help with annotation, the MBICS team for sharing their dataset as well as three anonymous reviewers for
feedback which improved our paper. All views are the authors' and do not necessarily reflect the views of any funders or affiliated institutions.}}

\titlerunning{Experiments in News Bias Detection}

\author{Tim Menzner\inst{1} \and Jochen L.~Leidner\inst{1,2}\orcidID{0000-0002-1219-4696}} %
\authorrunning{T.~Menzner and J.~L.~Leidner}


\institute{Information Access Research Group, Coburg University of Applied Sciences, Friedrich-Streib-Straße 2, 96459 Coburg, Germany \and University of Sheffield, Department of Computer Science, Regents Court, 211 Portobello, Sheffield S1 4DP, United Kingdom\\(Contact: \email{leidner@acm.org})}
\maketitle              
\begin{abstract}
  The World Wide Web provides unrivalled access to information globally, including factual news reporting and commentary. However, state actors and commercial players increasingly spread biased (distorted) or fake (non-factual) information to promote their agendas.

  We compare several large, pre-trained language models
  on the task of sentence-level news bias detection and sub-type classification, providing quantitative and qualitative results.
 
  Our findings are to be seen as part of a wider effort towards realizing the conceptual vision, articulated by Fuhr et al. \cite{Fuhr-etal:2018:SIGIRForum}, of a ``nutrition label'' for online content for the social good.

  \keywords{media bias \and propaganda detection \and content quality \and content nutrition label \and text analysis \and metadata enrichment \and pre-trained language models \and natural language processing \and information retrieval}
\end{abstract}


\section{Introduction} \label{sec:intro}

The triad of media bias, propaganda and fake news is
threatening democracy: the media, which have been
called the ``fourth estate'' of a democratic system,
are needed to inform citizens of the events of the
world, of the response of politicians that they
elected, and of potential scandals to keep the
political system honest.
Democracy may even be seen as facing an existential 
threat in scenarios where most citizens do not 
consume news about daily events from reliable 
sources, but from online commercial Websites or 
social media platforms whose business model are more 
aligned with propagating controversies than balanced 
news, because such a behavior promotes attention, 
and in an attention society, increased attention and 
stay time translates to higher advertising revenues.
Consequently, in response substantial research has
been conducted to -- manually or automatically --
determine the bias
of news publishers, publications or individual news
articles, story verification and fake news
detection, and spotting signs of propaganda.
Our own project \href{https://biasscanner.org}{biasscanner.org} aims to add to these efforts.

\textbf{Contribution.} In line with the social importance of this agenda, we present the first experiments to evaluate and compare the efficacy of a range of the latest
generation of pre-trained language
models, namely OpenAI GPT-3, GPT-4 and Meta Llama2, on the task of sentence-level news bias detection and sub-type classification, in both zero-shot and some fine-tuned settings. 

Any evidence obtained by these models can
then be used by search engines as part of their
ranking model, or post-hoc by users to filter out
highly biased material upon user
request.\footnote{
  We do not believe in automatic censorship. Citizens should be able to see bias material on request.
}


\section{Related Work} \label{sec:related-work}

\subsection{News Bias: The Phenomenon}

Media bias \cite{Lee-Solomon:1990,Groseclose-Milyo:2005:QuartJEcon,Sloan-Mackay:2007,Groeling:2016:AnnuRevPolitSci} has a
long tradition of being investigated, and often,
readers are aware of the political leanings of 
a newspaper, online or in paper form. However, the
increasing use of the Internet as a target for
information warfare and propaganda has led to an
increase of various types of bias, fake news and
other undesirable phenomena.
Evidence also suggests \cite{Vosoughi-Roy-Aral:2018:Sci} that fake news spreads
faster than balanced news.
DellaVigna et al. \cite{DellaVigna:2007} focused on the effect of news bias and voting patterns, suggesting a significant increase in republican votes in towns where Fox News entered the cable market in 2000. 
Groeling \cite{Groeling:2016:AnnuRevPolitSci} presents a survey of the literature covering partisan bias.
However, studying political bias in U.S. news reporting,
Budak et al. \cite{Budak-etal:2016:PublOpinQuart} found that ``news outlets are considerably more similar than generally believed. Specifically, with the exception of political scandals, major news organizations present topics in a largely nonpartisan manner, casting neither Democrats nor Republicans in a particularly favorable or unfavorable light.'' 

\subsection{Automatic Detection of News Bias}

After early pioneering work on bias from economics \cite{Groseclose-Milyo:2005:QuartJEcon}, Arapakis et al. \cite{Arapakis-etal:2016:ACL} labeled 561 articles along 14 quality dimensions including subjectivity.
Yano, Resnik and Smith \cite{Yano-Resnik-Smith:2010:NAACLWS} 
manually annotated sentence-level partisanship bias.
A system to detect sentiment in news articles was developed by Zhang et al. \cite{Zhang-etal:2011:HICSS} while Baumer et al. \cite{Baumer-etal:2015:NAACL} focused on detecting framing language. 
Bhuiyan et al. \cite{Bhuiyan-etal:2020:HCI} compare crowdsourced and expert assessment criteria for credibility on statements about climate change. 
Chen et al. \cite{Chen-etal:2020:FindEMNLP} demonstrated that incorporating second-order information, such as the probability distributions of the frequency, positions, and sequential order of sentence-level bias, can enhance the effectiveness of article-level bias detection, especially in cases where relying solely on individual words or sentences is insufficient.
Recasens et al. \cite{Recasens-etal:2013:ACL} exploit the Wikipedia edit history
based on ``violations of neutral point of view'' to obtain training
data, and theypresent a logistic regression model based on a lexicon
and various linguistic features.
Lim, Jatowk and Yoshikawa \cite{Lim-Jatowt-Yoshikawa:2020:DEIMForum} use a method based on 16 simple features to identify
bias-carrying terms: they first cluster news articles by their words at the document level, identifying similar event stories
as belonging to the same cluster, and then find contrasting paragraph/sentence pairs, one from insider and one from outside
a given cluster in the hope of obtaining representants of two quite different perspectives. Lim and colleagues are also to be credited for
first exploiting the value of rare (= high-IDF \cite{SparckJones:1972:JDoc}) words as bias-bearing.

Media bias datasets with different focus where
released by various groups \cite{Arapakis-etal:2016:ACL,Horne-etal:2018:ArXiv,Spinde-etal:2021:IPM,Spinde-Hamborg-Gipp:2020:ECMLPKDD}. 
MBIC, a media bias identification corpus, which we also use here, was introduced by Spinde et al.
\cite{Spinde2021MBIC}. 
Horne et al. \cite{Horne-etal:2018:ArXiv} released a larger dataset annotated for political partisanship bias, but without grouping articles by event, which makes apples-to-apples comparison harder; Chen et al. \cite{Chen-etal:2018:ICNLG} addressed this issue by resorting to another corpus sampled from the website \url{allsides.com}, which includes human labels by U.S. political orientation (on the ordinal scale $\{LL, L, C, R, RR\}$); they also present an ML model to flip the orientation to the opposite one.
Spinde et al. published a dataset containing biased sentences and evaluated detection techniques on it \cite{Spinde-etal:2021:IPM,Spinde-Hamborg-Gipp:2020:ECMLPKDD}.
Hube and Fetahu \cite{Hube-Fetahu:2018:WWW} were the first to
extract biased words from Wikipedia. They used crowdsourcing to
have humans annotate bias instances with a low reported Fleiss
$\kappa=0.35$ inter-coder agreement, suggesting the high subjectivity
of the task as defined.
Conrad et al. \cite{Conrad-Leidner-Schilder:2008:WICOWWS} focused on content mining to measure
credibility of authors on the Web; credibility is orthogonal but
related to bias. Ghanem et al. \cite{Ghanem-etal:2021:EACL} analyze 
an interesting way to distinguish between real/credible news and fake
news by looking at the distribution of affective words within the
document.
Spinde et al. \cite{Spinde-etal:2021:IPM}
tried to extract a lexicon of bias workds from ordinary news. 
The Ph.D. thesis of Hamborg \cite{Hamborg:2023} 
provides a frame-oriented bias analysis technique.
Hamborg et al. \cite{Hamborg-etal:2020:JCDL} provided a recent and
interdisciplinary literature review to suggest methods how bias could
be bias detection could be automated. See \cite{Augenstein:2021:DSci}
for a good survey of the related area of automated and explainable
fact checking from an NLP perspective.

Our work is perhaps closest in spirit to that of Wessel et al. \cite{Wessel:2023:online}, who evaluated transformer techniques on
detecting nine different types of bias across 22 selected datasets, which they obtained and merged
(= their MBIB dataset). In contrast to our work,
that paper used a different dataset 
(MBIB$\neq$MBIC) and they did not compare to the
any of the transformer types we used, so their
findings are not directly comparable.
To the best of our knowledge, no previous work has compared OpenAI GPT-3.5, GPT-4 and Meta Llama2
with each other or a fine-tuned version of one of
them, for the news media bias detection task.


\section{Data} \label{sec:data}

We evaluated using the MBIC dataset \cite{Spinde2021MBIC}, which consists of 1,700 statements that contain various occurrences of media bias, annotated by 10 different judges. 
The statements were taken from eight different US-American news outlets: The HuffPost, MSNBC, and AlterNet as examples of left-oriented media; The Federalist, Fox News, and Breitbart to represent right-wing media outlets; and two sources from the center: USA Today and the Reuters news agency.  

After removing those statements for which annotators could not arrive at a final decision, 1,551 statements (1,018 biased, 533 non-biased) were left for us to use. 


\section{Methods} \label{sec:method}

We conducted three lines of experiments to assess quantitatively and qualitatively the ability of
three of the latest-generation large pre-trained
languge models, GPT-3.5, GPT-4 and Llama2 (Table \ref{tab:methods}), when
employed to classify sentences from English news with respect to whether or not they may express
any type of bias, and if so, which type.
\vspace{-4mm}
\begin{table}
  \caption{Methods Used in Our Experiments}
  \label{tab:methods}
  \begin{tabular}{lll}\hline
    \textbf{Method 1}. & Open AI, GPT-3.5 & one/zero-shot + prompt engineering, called via API \\

    \textbf{Method 2.} & Open AI, GPT-3.5 & prompt engineering, fine-tuned by us, via API \\

    \textbf{Method 3.} & Open AI, GPT-4 & one-shot, via API \\

    \textbf{Method 4.} & Meta Llama2 & one-shot, on premise  \\ \hline
  \end{tabular}
\end{table}
\vspace{-3mm}
%
%
All of these were evaluated using MBIC. When iterating through the dataset for evaluation, two different modes were used. 
In the first mode, sentences were not evaluated individually, but in batches joined together in groups of ten sentences, separated by a line break.
This was done for the sole benefit of minimizing the number of API calls, as the model had not to be prompted for each single sentence. 
Also, this might be considered a more natural case for bias detection, as it models the process of bias identification in longer texts. 
As the cleaned dataset contained 1,551 sentences, which can not be divided by 10 without a remainder, the last sentence was removed from the dataset in this mode. The results for this evaluation can be found in Table \ref{table:classification_results_10_sent_blocks} below (Section \ref{sec:eval}).

In the second mode (results shown in Table \ref{table:classification_results_individual_sentences} below), sentences were evaluated \emph{individually}, independent of preceding or following sentences, to see whether batching unrelated sentences in the first line of experiments may have harmed performance of the models.
The language model's instruction prompt underwent iterative development\footnote{
  The text of the various prompts used here is too long to include it in
  this paper; we will share them in the GitHub repository at (Anonymized) upon acceptance of this
  paper.
} to guarantee the production of consistent and high-quality results. It incorporates a precise delineation of each sought-after bias category (linguistic bias, text-level context bias, reporting-level context bias, cognitive bias, hate speech, fake news, racial bias, gender bias, and political bias), with the definitions given by \cite{Wessel:2023:online}, along with an illustrative example demonstrating the desired JSON output structure. 
Subsequently, the model's output underwent a post-processing and filtering phase to mitigate potential errors before flagging biased sentences within the content. 

For each bias-flagged sentence, the model was also asked to offer insights into the bias type, a detailed explanation, and a quantified bias strength score. 
The development of the prompt followed evaluated best practice techniques, like asking the model to assume an expert role, breaking the task into smaller sub tasks to complete step by step or providing an example output for one shot prompting \cite{OpenAI:2023:online}.
While the evaluations presented in Table \ref{table:classification_results_10_sent_blocks} and Table \ref{table:classification_results_individual_sentences}. were all conducted with the same prompt, several additional
variations were tested for the experiments presented in Table \ref{table:classification_results_variations_10_sent_blocks}, to gain insights in the effect of different prompt engineering techniques and parameter changes on the quality of results. 
All evaluations unless stated otherwise were conducted with a temperature of 0.0 to keep the model's outputs more consistent. 
Evaluations in Table \ref{table:classification_results_10_sent_blocks} and Table \ref{table:classification_results_individual_sentences} were conducted using GPT-3.5-turbo-16k, two different fine-tuned GPT-3.5-turbo models (called Variant A and B, respectively in the results tables below), GPT-4 and Llama2-70b-chat-awq. 
The first fine-tuned model, for the experiment presented in Table \ref{table:classification_results_10_sent_blocks}, was fine-tuned with 50 examples, each consisting of the system prompt, the article built from 10 sentences and a desired model output for this case (FT Variant A).
While the sentences considered biased could be taken directly from the dataset, the information about bias type and bias score (which was not relevant to the result of the evaluation, but was needed to stay in format), was filled in using GPT-4. 
The 500 sentences (50 batches a 10 sentences) used for fine-tuning were removed from the dataset before the evaluation of the model. 
For the experiments presented in Table \ref{table:classification_results_individual_sentences}, fine-tuning was conducted in a similar manner, but using 50 held-out examples comprising \emph{individual} sentences (FT Variant B).
GPT-3.5-turbo-16k served as the foundational model for all the tests in Table \ref{table:classification_results_variations_10_sent_blocks}. This decision allowed us to conveniently evaluate the effects of different prompt and parameter modifications. It is important to note that different models may produce varying responses to these changes. 

In the third line of experiments,
next to the described prompt, different variations were evaluated. 
One version where a (fictional) example given in the original prompt was removed, to see how significant the influence of such an example can be.
In a different approach, we excluded the definitions of text-level context bias, reporting-level context bias, and cognitive bias from the prompt. These concepts heavily rely on context, which the dataset we are using cannot provide, as explained in Section \ref{sec:data}. 
In another scenario, we completely omitted the explanation of different bias types. This was done to assess the model's performance when relying solely on its internal knowledge of bias. 
For yet another different experiment, we implemented a second round of prompt engineering techniques. This involved making subtle modifications to enhance clarity in instructing the model on the specific steps to follow, along with replacing certain words with more precise or commonly used expressions. 
In a separate run, we leveraged the model's bias score to filter out sentences in which only a mild bias was detected. Specifically, we removed all sentences with a bias score less than 0.6 from the results. 
For another experiment, we set no custom system prompt but instead appended the text that would have gone there at the start of the user message. While using the system message for prompting the basic behaviour of the model is considered best practice, there had initially been discussion about it not having the same weight as the user message \cite{openaisystemRole}.
Finally, in a final experiment, we set the model temperature from 0.0 to  GPT's default of 0.7, to test a variant that uses more influence of randomness.


\section{Evaluation} \label{sec:eval}

In this section, we report the results of our experimental findings.

\subsection{Quantitative Evaluation} 

Table \ref{table:classification_results_10_sent_blocks} shows the results for the comparison
of GPT-3.5, both \emph{as is}, as well as versus our fine-tuned variant of it, with GPT-4 and Llama2.
In this first set of experiments, we posed the news text to be classified
to the models in groups of ten sentences at a time without overlap.
Our fine-tuned version of GPT-3.5 outperformed all other models in terms of F1 through a
substantial increase of Precision by 14\%, traded for a drop in Recall by 16\%.
Overall, GPT-4 has the highest precision (84\%) in absolute terms, but its small lead of 2\% over
the fine-tuned GPT-3.5 is not worth the added energy, cost and memory for almost all applications, in
particular given that its Precision is also lower.

\begin{table}
\caption{Evaluation Results for GPT-3.5-turbo-16k,  GPT-3.5-turbo fine-tuned, GPT-4 and Llama2-70b-chat-awq, on 10 sentence blocks. Best results are highlighted in bold.}
\label{table:classification_results_10_sent_blocks}
\begin{tabular}{lrrrrrrr}
\toprule
\textbf{Model} & \textbf{TP} & \textbf{FP} & \textbf{FN} & \textbf{TN} & \textbf{F1-Score} & \textbf{Recall} & \textbf{Precision} \\
\midrule
  GPT-3.5                         & 965 & 460 &  53 &  72 & 0.790          & \textbf{0.948}  & 0.677 \\
  \textbf{GPT-3.5 (FT Variant A)} & 442 & 100 & 119 & 189 & \textbf{0.802} & 0.788           & 0.816 \\
  GPT-4                           & 739 & 138 & 279 & 394 & 0.780          & 0.726           & \textbf{0.843} \\
  Llama2                          & 579 & 241 & 439 & 291 & 0.630          & 0.569           & 0.706 \\
\bottomrule
\end{tabular}
\end{table}


Table \ref{table:classification_results_individual_sentences} shows the results for a different
set of experiments, in which sentences were posed individually to the models, leaving out
the formation of 10-sentence blocks, in order to assess the degree of influence the mixture of different biased and unbiased sentences in a longer text has on the transformer models' ability to make the right decisions.
The model's decision about the bias level of a sentence might not be absolute but relative to the bias exhibited by other sentences included in the same prompt. 
To our surprise, GPT-3.5 did
quite well on individual sentences, slightly better than on groups of sentences even, whereas fine-tuning appears to benefit from longer texts, since the performance of the model trained on single sentences, dropped 
substantially (-14\%). GPT-4 improved in the absence of other sentences, from 78\% to 82\% in terms of F1.

\begin{table}
\caption{Evaluation Results for GPT-3.5-turbo-16k, GPT-3.5-turbo fine-tuned, GPT-4 and Llama2-70b-chat-awq, on individual sentences. Best results are highlighted in bold.}
\label{table:classification_results_individual_sentences}
\begin{tabular}{lrrrrrrr}
\toprule
\textbf{Model} & \textbf{TP} & \textbf{FP} & \textbf{FN} & \textbf{TN} & \textbf{F1-Score} & \textbf{Recall} & \textbf{Precision} \\
\midrule
  GPT-3.5                &  943 & 370 &  75 & 163 & 0.809 &  0.926 & 0.718 \\
  GPT-3.5 (FT Variant B) &  527 &  82 & 455 & 437 & 0.662 &  0.537 & 0.865 \\
  GPT-4                  & 1003 & 415 &  15 & 118 & 0.823 &  0.985 & 0.707 \\
  Llama2                 & 1014 & 525 &   2 &  10 & 0.793 & \textbf{0.998} & 0.659 \\
\bottomrule
\end{tabular}
\end{table}

In addition, we conducted an extra set of experiments with GPT-3.5,
in which  we varied prompts and parameter settings. The results are
shown in Table \ref{table:classification_results_variations_10_sent_blocks}.

As these results show, providing an example within the prompt only had a minor effect on the outcome, whereas removing the definitions for context dependent bias criteria led to a small shift towards fewer total positives and slightly more negatives, while keeping the F1-Score consistent. 
Restructuring the prompt to substitute unclear words and to work out the division into individual work steps more clearly, led to an increased F1-score and small gains in precision, while the recall performance dropped at the same time. 
As expected, filtering out results where the bias score was below 0.6, led to a notable increase in precision and a notable decrease in recall. 
Interestingly, giving the model no definition of bias and different bias types at all, resulted in the best F1-score, with a higher precision and a lower recall. 
In this setting, the model had to come up with its own definition of bias types. Most of the times, it selected a generic categorisation like ``negative bias'' (21\%), ``positive bias'' (4\%) or just ``bias'' (15\%). 
However, it also identified more specific bias categories like ``political bias'' (15\%), ``loaded language'' (12\%), ``spin'' (7\%), ``emotional bias'' (2\%), ``gender bias'' (2\%) or ``bias by omission'' (1\%).
Some of these are identical or resemble the categories used in our prompt.
In many cases, the model further appears to use different spellings or different wording potentially describing the same thing like ``omission'' (1\%) additionally to the already mentioned ``bias by omission'' or ``sexism'' (0.4\%) next to the discussed ``gender bias''.
The model also came up with rather obscure and specific bias categories like ``bias against conspiracy theories'' or ``bias against Bernie Sanders'', respectively exactly one time.

In comparison, our fine tuned GPT 3.5 model, which was the best performing model in the 10 batch mode, managed to stick to the defined bias categories nearly every time, spread as following: Political bias (57\%), linguistic bias (13\%), reporting-level context bias (8\%), text-level context bias (8\%), gender bias (5\%), racial bias (4\%), fake news (0.2\%). The remaining percentages were split on cases where the model named two different biases at the same time (3\%) or introduced own categories like ``economic bias'' or ``generational bias'', which could also all be found in the results of the run without definitions in the prompt  (1\%).  

Furthermore, using a larger user message instead of submitting a non-changing basic instruction as a system message, which is considered best practice, might increase recall and decrease precision.
Ultimately, setting the temperature of the model back to the GPT default of 0.7 up from the otherwise used 0.0, decreased the performance on all evaluated metrics. 
While these tests indicate that certain modifications to prompt and technique can have an influence on the performance, it is important to bear in mind that the non deterministic nature of large language models might influence the results in one or the other direction. 

\begin{table}
\caption{Evaluation results for GPT-3.5-turbo-16k with different prompt variants and parameter settings, on 10 sentence blocks. Best results are highlighted in bold.}
\label{table:classification_results_variations_10_sent_blocks}
\begin{tabular}{lrrrrrrr}
\toprule
\textbf{Model} & \textbf{TP} & \textbf{FP} & \textbf{FN} & \textbf{TN} & \textbf{F1-Score} & \textbf{Recall} & \textbf{Precision} \\
\midrule
  GPT-3.5 Base            & 965 & 460 &  53 &  72 & 0.790 \ud{} & 0.948 \ud{} & 0.677 \ud{} \\
  No Example in Prompt    & 966 & 468 &  52 &  64 & 0.788 \dn{} & \textbf{0.949} \up{} & 0.674 \dn{} \\
  No Contextual/Cognitive & 938 & 426 &  80 & 106 & 0.788 \dn{} & 0.921 \dn{} & 0.688 \up{} \\
  Restructured Prompt     & 956 & 439 &  62 &  93 & 0.792 \up{} & 0.939 \dn{} & 0.685 \up{} \\
  Bias Score Threshold    & 632 & 241 & 273 & 110 & 0.711 \dn{} & 0.698 \dn{} & \textbf{0.724} \up{} \\
  No Bias Definition      & 894 & 344 & 124 & 188 & \textbf{0.793} \up{} & 0.878 \dn{} & 0.722 \up{} \\
   No System Prompt   & 986 & 486 & 32 &  46 & 0.792 \up{} & 0.969 \up{} & 0.670 \dn{} \\
  High Temperature        & 931 & 446 & 87 &  86 & 0.778 \dn{} & 0.915 \dn{} & 0.676 \dn{} \\ \bottomrule
\end{tabular}
\end{table}

All of these numbers above are related to the binary +/-BIAS
sentence-level classification, for which MBIC contained the
binary gold labels. Our modes also output sub-categories,
for which no gold label was available, so we drew a random
sample of $N=100$ sentences. Three  human annotators
($K=3$) independently judged the model output either as
``right'' or ``wrong''. The result, per category and overall,
is shown in Table \ref{table:subcategory-eval}.
%
Using automatic creation of a silver dataset via majority
voting between the three human judge's decision, the overall accuracy
of the model was $A=87\%$.

\begin{table}
\begin{small}
  \caption{Evaluation Results for the Bias Subcategory Evaluation GPT 3.5 (FT Variant A) model (random sample $N=100, K=3$, silver label through majority voting)}
  \label{table:subcategory-eval}
  \begin{tabular}{lrrrr}\hline
    \textbf{Bias Sub-Type}             & \textbf{Correct} & \textbf{Incorrect} & \textbf{Total} & \textbf{Accuracy} \\ \hline
    Political bias                     &      56 &         7 &    63 &  88.89\% \\
    Racial    bias                     &       5 &         1 &     6 &  83.33\% \\
    Gender bias                        &       3 &         1 &     4 &  75.00\% \\
    Text-level context bias            &       5 &         1 &     6 &  83.33\% \\
    Linguistic bias                    &      11 &         1 &    12 &  91.67\% \\
    Reporting-level context bias       &       7 &         0 &     7 & 100.00\% \\ 
    Sub-type violation (hallucination) &     n/a &         2 &     2 &      n/a \\ \hline
    Total (all classes)                &      87 &        13 &   100 &  87.00\% \\ \hline
  \end{tabular}
\end{small}
\end{table}
%

\vspace{-0.5cm}

\subsection{Qualitative Evaluation}  \label{sec:qualitative}

Overall, we are satisfied with the quality pre-trained transformers
can detect instances of news bias on \emph{sentence granularity} and
classify them into sub-categories. 
Some examples of what could be considered a high quality output, are presented in Example \ref{ex:1}.
We are less satisfied with the level of our
understanding of some sub-classes of news bias and
the degree of subjectivity that is still
visible from the various gold data sets available
today. Examples for questionable model assessments
can be found in Example \ref{ex:2}.

\begin{table}
\begin{example} \label{ex:1} Examples for high quality model decisions\\
   \begin{small}
     \begin{tabular}{{p{0.25\linewidth}  p{0.7\linewidth}}} \hline
       sentence text & In October, Breitbart London reported that French police are allegedly actively assisting migrants to avoid detection by UK Border Force, telling them when and when not to attempt to cross the channel, according to a people smuggler. \\
      gold label & Biased \\
      assessment by model & GPT-3.5 FT \\
      bias\_type & Reporting-level context bias \\
      bias\_score & 0.5 \\
      bias\_description & The sentence cites Breitbart London as the source of the information, which is known for its conservative bias. This could be seen as a bias in the choice of source, potentially influencing the reader\'s perception of the reported event. \\ \hline
      sentence text & Football supports Trump in its promotion of racial division, the crushing of dissent, and the spread of misinformation, inequality, and brutality. \\
      dataset label & Biased \\
      assessment by model & Llama2 \\
      bias\_type & Political bias \\
      bias\_score & 0.9 \\
      bias\_description & This sentence directly implies that football, by extension the NFL, supports Trump and his policies, which is a highly controversial and debated topic. The use of the phrase 'promotion of racial division' and 'spread of misinformation' implies a strong negative bias against Trump and the NFL, and could be seen as taking a political stance. \\ \hline
      sentence text & On one hand, naive teenagers who signed off their financial futures to leftist, anti-American institutions would catch a break, The college cartel screwed you; now here’s a government waiver to make you forever grateful to the Democrats. \\
      gold label & Biased \\
      assessment by model & GPT-4  \\
      bias\_type & Linguistic bias \\
      bias\_score & 0.8 \\
      bias\_description & The use of the term 'naive teenagers' and 'leftist, anti-American institutions' introduces a negative judgement and bias against those who support college debt forgiveness and the institutions that provide higher education. The sentence also implies that the Democrats' motive for the policy is to make people 'forever grateful' to them, suggesting manipulation. \\ \hline
      sentence text & Male and female are the two sexes — but transgender advocates now claim more than 50 “genders” — including agender, bigender, gender fluid, gender non-conforming, genderqueer, intersex, two-spirit, etc, plus “cis-gender” for the 99+ percent of people who are normal, non-transgender people. \\
      gold label & Biased \\
      assessment by model & GPT-4 \\
      bias\_type & Gender bias \\
      bias\_score & 0.7 \\
      bias\_description & The sentence uses the term 'normal' to describe non-transgender people, implying that transgender people are not normal, which is a negative portrayal \\ \hline
    \end{tabular}
  \end{small}
\end{example}
\end{table}

\begin{table}
\begin{example} \label{ex:2} Examples for arguably questionable model decisions



   \begin{small}
     \begin{tabular}{{p{0.25\linewidth}  p{0.7\linewidth}}} \hline
       sentence text &  A study published Tuesday in the open peer-reviewed journal Proceedings of the Royal Society B finds that the microorganisms in the wax worm's gut help them consume and metabolize plastics. \\
      gold label & Non-biased \\
      assessment by model & Llama2 \\
      bias\_type & Reporting-level context bias \\
      bias\_score & 0.5 \\
      bias\_description & The article focuses on a specific study and its findings, without providing a broader context or mentioning other research on the topic. This may create a bias towards the study's conclusions and findings, without considering alternative viewpoints or research. \\ \hline
      sentence text & A Gallup poll reports that women are still more likely to do laundry, clean the house, do grocery shopping, prepare meals, wash dishes and make decisions about furniture and decorations – even among younger generations who are reportedly more egalitarian as ever. \\
      gold label & Non-biased \\
      assessment by model & GPT-3.5 \\
      bias\_type & Gender bias \\
      bias\_score & 0.9 \\
      bias\_description & This sentence reinforces traditional gender roles by stating that women are more likely to perform household chores and make decisions about furniture and decorations. It implies that these tasks are inherently feminine and perpetuates gender stereotypes. \\ \hline
      sentence text & 'Hundreds of white nationalists lined the streets of Charlottesville, Va., last Saturday to protest the removal of a Confederate monument, Some waved Confederate and Nazi flags, others brandished shields \\
      gold label & Non-biased \\
      assessment by model & GPT-3.5-FT \\
      bias\_type & Reporting-level context bias \\
      bias\_score & 0.7 \\
      bias\_description & The sentence focuses on the presence of white nationalists and their use of controversial symbols, which could be seen as a negative portrayal of this group. \\ \hline
    \end{tabular}
  \end{small}
\end{example}
\end{table}

As visible in Example \ref{ex:1}, the model can also utilize the names of certain media outlets, renowned for their perceived bias, as an indicative factor for assessing bias.
When reporting merely factual statements, the models sometimes judged those as ``biased'', asking for an unreasonable amount of potentially unnecessary
context. The mentioning of facts and findings about existing biases in our society was also occasionally flagged as biased.  
Similarly, this tendency persisted when reporting on actions that could be considered criminal or unethical, of individuals or groups; even when such reports were objectively presented, the models occasionally accused the articles of introducing a negative bias against the subject. \emph{We conjecture that there is an underlying inability of the models to distinguish between language and meta-language, and to handle reported speech.}
Given that all models were trained on a fixed dataset and do not automatically get updated with more recent data, their ability to detect fake news in contemporary articles is constrained, suggesting that the ``fake news'' bias category may not be of use in this context. As the evaluation dataset did not contain instances of fake news, this was not a problem for the evaluation. However, similar to the phenomena described above, a statement mentioning fake could get marked with ``fake news'' as bias type, just for mentioning it.

When running the evaluation on individual sentences, we also observed that sometimes, the models hallucinated more sentences based on the actual sentences from the dataset or they contradicted themselves in its assessment of the sentence and their overall conclusion about the "article". 
This did not influence the results of the evaluation as only the first sentence was taken into account, but it serves as evidence of the potential instability in the output of large language models.
Instability, which could also be noted by occasional syntax errors in the JSON, which the models were asked to provide. 
In sporadic cases, the model further included non biased sentences with a bias type like "None" and a bias score of 0 in its assessment, which were filtered out before the analysis.
While this filter was created with the intention of capturing as many cases as possible, but due to the inherent unpredictability of large language models, a 100\% coverage can not be guaranteed.
As described in Section \ref{sec:limitations}, the dataset also contained instances of debatable annotation, as shown in Example \ref{ex:3}, for instance:


\begin{table}
\begin{example} \label{ex:3} Examples for arguably debatable model decisions

   \begin{small}
     \begin{tabular}{{p{0.25\linewidth}  p{0.7\linewidth}}} \hline
       sentence text &  But the elimination of air travel strikes particularly close to the homes of Sens, Kirsten Gillibrand, Cory Booker, Bernie Sanders, and Kamala Harris – all of whom extensively used air travel, including private jets – throughout the years in office. \\
      gold label & Non-biased \\
      assessment by model & GPT-3.5 \\
      bias\_type & Reporting-level context bias \\
      bias\_score & 0.8 \\
      bias\_description & The sentence highlights the air travel habits of specific senators who support the elimination of air travel, suggesting hypocrisy or inconsistency in their actions. \\ \hline
      sentence text & Under current immigration policies, it’s possible that immigration arrests could lead to poor mental health by increasing deportation fears among undocumented individuals and their families and neighbors, the study authors write. \\
      gold label & Biased \\
      assessment by model & GPT-3.5-FT \\
      bias\_type & None \\
      bias\_score & 0 \\
      bias\_description & The sentence does not exhibit any notable bias. \\ \hline
    \end{tabular}
  \end{small}
\end{example}
\end{table}

\section{Limitations and Ethical Considerations}
\label{sec:limitations}

No automatic neural or statistical classifier is capable of detecting
instances of news bias perfectly; a certain number of false positives
and false negatives are bound to occur. 
A comparison with other automated approaches was out of scope for this paper, as we found the techniques hard to replicate and the circumstances of their evaluations were different. 
Several forms of bias depend on \emph{context}. This particularly affects text-level context bias, reporting-level context bias and cognitive bias. With the dataset consisting of single sentences instead of a paragraph or whole articles, the necessary context is missing that would be needed to take the right decision.
Residual controversial annotations make
the evaluation difficult. For instance, when one language model  identifies a sentence as biased but the data set does not list it as such, this might be logged as a false positive, sometimes in error (see
Example \ref{sec:qualitative}). 
Furthermore, since the MBIC dataset only contains news from U.S. outlets, it is inherently strongly biased towards topics like U.S. politics in
the period of the collection of the data (Trump, abortion).
The fact that all annotators were based in the U.S. only adds to the U.S.-centric focus. Perceptions of bias can vary across different cultures and may also evolve over time within the U.S itself. 
Considering that the MBIC Dataset was published early enough to be incorporated into the training data of all models, it remains a possibility that certain parts of it were included, even though none of the three models used were familiar with a dataset by that name when queried.

Bias detection has its own biases: not all bias types are recognized
equally well. To mitigate risks pertaining to this, we urge
application developers that use our or other methods to make the users
of their end applications aware of this. 
Furthermore, when dealing with the question of news bias, the topic of
\emph{false balance} has to be considered. While presenting different perspectives on an issue could be described as an unbiased approach, false balance means giving equal weight or airtime to views that are not supported by credible evidence, which can again create bias in form of a misleading perception of a balanced presentation. This phenomena has been extensivly studied in the context of climate change research, where false balance has been shown to influence public perceptions and beliefs \cite{imundo2022fairness}.
When analysing reporting using negatively connotated words to describe someone's position as ``racist'', for example, can be either unjustified
biased or an objective description of this person's ideology.
Avoiding such negative connotated words when they would provide an accurate depiction could, in itself, introduce a form of bias.
Furthermore, the potential instability and unpredictability of the output of large language models (hallucination) poses a challenge for building applications.
Finally, the use of models over a networked API (as in the case of GPT-3.5 and GPT-4) further poses potential privacy risks as content to be transferred via an API must be shared with the server-owning counterparty, or
risk of unavailability in case of network downtime.


\section{Summary, Conclusions and Future Work} \label{sec:conclusion}

In this paper, we described a set of experiments using several
pre-trained neural transformers to compare their performance on the
task of news bias detection on multiple datasets. To have a capability
for high-quality automatic news bias identification and classification
is socially desirable in order to avoid that users of the Web drown in
propaganda, biased and fake news ``reporting'', so we view our
contribution as much as a social one as a scientific one.

Our experiments show that a fine-tuned variant of a model with the
smaller number of parameters (GPT-3.5) can outperform a model with a
much larger number of parameters (GPT-4). 
We also discovered in our analysis that all models struggled with reported speech, with distinguishing language and meta-language, and with hallucinating ``new'' uncalled-for categories, all of which should be explored further.

\sloppypar{}
In future work, we plan to work on cross-distilling a single open,
non-proprietary, on-premise model that is fine-tuned on the media bias
identification and classification task, and on rolling out a browser
plug-in (Anonymized) in order to put a tool into citizens' hands to
promote their critical reading of the news, in the spirit of \cite{Fuhr-etal:2018:SIGIRForum}. 
We also plan to collect a larger corpus of biased news specimens in
multiple languages, and invite other research groups to join forces on
this socially relevant endeavour.

%



\bibliographystyle{splncs04}
\bibliography{news-bias}


\end{document}